# Using Random Effects Machine Learning Algorithms to Identify Vulnerability to Depression

Runa Bhaumik[1], Jonathan Stange[2]

Author affiliations

[1]University of Illinois at Chicago, Department of Psychiatry
[2]University of Southern California

Email: rbhaumik@uic.edu

   jstange@usc.edu

 *  Corresponding author: rbhaumik@uic.edu

**Abstract**

**Background:**

Reliable prediction of clinical progression over time can improve the outcomes of depression. Little work has been done integrating various risk factors for depression, to determine the combinations of factors with the greatest utility for identifying which individuals are at the greatest risk.

**Method:**

This study demonstrates that data-driven machine learning (ML) methods such as RE-EM (Random Effects/Expectation Maximization) trees and MERF (Mixed Effects Random Forest) can be applied to reliably identify variables that have the greatest utility for classifying subgroups at greatest risk for depression. 185 young adults completed measures of depression risk, including rumination, worry, negative cognitive styles, cognitive and coping flexibilities, and negative life events, along with symptoms of depression. We trained RE-EM trees and MERF algorithms and compared them to traditional linear mixed models (LMMs) predicting depressive symptoms prospectively and concurrently with cross-validation.

**Results:**

Our results indicated that the RE-EM tree and MERF methods model complex interactions, identify subgroups of individuals and predict depression severity comparable to LMM. Further, machine learning models determined that brooding, negative life events, negative cognitive styles, and perceived control were the most relevant predictors of future depression levels.

**Conclusions:**

Random effects machine learning models have the potential for high clinical utility and can be leveraged for interventions to reduce vulnerability to depression.

**Introduction**

Major depressive disorder (MDD) is a prevalent and debilitating disorder and is associated with tremendous personal and societal costs (Kessler & Wang, 2009). It is also one of the most common mental disorders among college students (Click or tap here to enter text. Auerbach et al., 2016, 2018; Farabaugh et al., 2012). Research has indicated that depression in college students is associated with lower academic performance (Hysenbegasi et al. 2005), increased levels of anxiety (Rawson et al. 1994), alcohol and drug dependency, poorer quality of life and self-harming behaviors (Serras et al. 2010). Thus, given the high prevalence and burden of MDD, more research is needed to evaluate such risk factors for the development of depressive symptoms. Hypothesis-driven studies typically predict only limited variance in depression. To better identify depression, there is a need to improve our models to predict who is at the greatest risk, and when that risk is most likely to increase.

Flexibility – the ability to adapt thoughts and behaviors to meet changes in contextual demands – has been proposed as a group of broadly-related constructs that may serve as protective factors against depression (Kashdan et al. 2010; Stange et al., 2016, 2017). Several approaches to assessing flexibility, or the lack thereof (inflexibility), have been evaluated as predictors of depressed mood, including coping flexibility, explanatory flexibility, perseverative thinking such as rumination and worry, and cognitive flexibility. Other known risk factors for depression include exposure to negative life events (Hammen, 2005), having a negative cognitive style of responding to negative events (Alloy et al., 2017), and engaging in maladaptive emotion regulation strategies (Aldao et al., 2010; Dryman & Heimberg, 2018). However, research typically has examined one or two of these constructs in isolation; little work has integrated these risk factors or investigated them comparatively as predictors of depression longitudinally (Stange et al., 2017). The goal of this study is to highlight machine learning techniques that can facilitate the identification of risk and protective factors, and their interactions, which may reduce vulnerability to developing symptoms of depression in young adults. Although the group of risk factors we consider is not exhaustive, we present this analysis as an example of how machine learning techniques can be used to integrate diverse measures in the evaluation of risk, in ways that may have clinical utility.

**When and How Might Random-Effects Machine Learning Modeling Have Utility for Clinical Scientists?**

Statistical analysis of longitudinal data requires accounting for possible between-subject heterogeneity and within-subject correlation. Linear Mixed effects (LMM or multilevel) modeling is one of the approaches [Laird et al. 1982, Liang et al. 1986] that is typically used to analyze longitudinal data accounting for the correlated nature of data. However, there are several limitations in traditional mixed effects modeling, including parametric assumption, labor-intensive testing for complex interactions, and overfitting due to a large number of attributes.

Random effects machine learning modeling may be useful in situations common to health science researchers. This is particularly true for heterogeneous disorders such as depression (Goldberg, 2011; Monroe & Anderson, 2015), for which there are several proposed risk or protective factors, the outcome variable is determined by several risk factors and the complexity of the interaction of these factors to determine an individual's likelihood of this outcome behavior.

Regression trees can identify subgroups of individuals who share common characteristics and produce a user-friendly visual output (which is referred to as a decision tree). This output then can be provided to clinicians to tailor further assessment and intervention. One of the advantages of regression tree methods is that they do not require a hypothesized pattern of association between the predictors and the outcome. Moreover, higher-order interactions, and thus non-linear associations between the predictors and outcome variables, are allowed via the different splits of nodes in the regression tree. From a clinical perspective, this data-driven approach is a clear benefit. Examples of relevance to depression research include the detection of moderators. For example, negative life events are well established as one of the strongest predictors of depression, and a variety of vulnerability factors have been identified that moderate the link between life events and depression (Alloy et al., 2017), such as negative or inflexible cognitive styles, and interpersonal and emotion-regulatory vulnerabilities. However, for reasons of power and collinearity, it often is infeasible to include more than a few potential moderators simultaneously in mixed effects models, even though numerous potential moderators exist. Regression tree methods can help to address this critical limitation of the models typically used in depression research.

**Aim of the Present Study**

The current paper aims to introduce both decision tree and random forest methods that allow for the analysis of multilevel and longitudinal datasets: random effects regression trees (Random Effects/Expectation Maximization (RE-EM) trees, Sela & Simonoff, 2012) and mixed effects random forest (MERF, Hajjem et al 2014). The goal of this study is to explore whether tree-based machine learning algorithms can be used as forecasting tools to identify young adults who are at risk of developing depression in the future. This study addresses the following research questions:

1) How well do random effects machine learning algorithms perform in terms of predicting depression severity, relative to traditional LMMs?
2) What combination of risk and protective factors is best at identifying which individuals are most likely to have *concurrent* symptoms of depression? For example, do people with inflexibility and maladaptive regulation have higher depression severity than those who are more flexible and use adaptive regulation strategies?

**Method**

**Participants and procedure**

Participants were undergraduate students at a large urban university, recruited using flyers on campus and from undergraduate psychology classes. To be included in the study, participants were required to have normal or corrected-to-normal vision and be fluent in English. Students received psychology course credit or were compensated in cash for participation. All participants provided written informed consent approved by the University's Institutional Review Board. The sample included 185 participants (56.7% female, 66.2% Caucasian/White, 19.1% African American/Black, 13.3% Asian/Pacific Islander, 0.6% Native American, 5.1% other race, and 8.3% Hispanic/Latino), with a mean age of 21.98 years (SD = 5.90).

Participants completed a set of self-report questionnaires at baseline, and at four follow-up assessments (Times 2-5) spaced three weeks apart. This time frame was chosen to allow for sufficient variability in negative life events while enabling the modeling of short-term fluctuations in internalizing symptoms (Stange et al., 2017). All questionnaires used in the present analyses were administered at baseline. During the follow-up assessments, participants completed measures of rumination, worry, anxiety, depressive symptoms, and exposure to negative life events in the prior three weeks. At Time 5, participants were interviewed to verify that life events reported met *a priori* criteria and occurred within the correct three-week interval.

Participants were required to have completed at least one of the four follow-up assessments to be included in the present analyses, which yielded a final sample size of 185. The measures included in our research are described in Supplementary Materials Section A. The measures are the Cognitive Flexibility Inventory, Coping Flexibility (Scale and Questionnaire), Emotion Regulation Questionnaire, Life Events Scale and Interview, Penn State Worry Questionnaire, Ruminative Response Scale, and Cognitive Responses to Life Events. The summary of measures and variables used in the models is shown in Table 1.

**Table 1**. Summary of measures and variables used in models.

| Measures | Variables used in the model |
|---|---|
| Cognitive Flexibility Inventory | Perceived alternatives |
|  | Perceived controllability |
| Coping Flexibility Scale | Problem-focused coping flexibility |
|  | Emotion-focused coping flexibility |
| Coping Flexibility Questionnaire | Coping flexibility |
| Emotion Regulation Questionnaire | Reappraisal |
|  | Suppression |
| Life Events Scale and Interview | Negative life events |
| Penn State Worry Questionnaire | Worry |
| Ruminative Response Scale | Brooding |
|  | Pondering |
| Cognitive Responses to Life Events | Negative cognitive styles |

**Statistical and Machine Learning Algorithms**

In this work, we empirically assessed the performance of a linear mixed effect model, RE-EM tree, and a mixed effects random forest (MERF) method on repeated measurements of depression severity. Mixed effects models are used in statistics and econometrics for longitudinal data, where observations are collected multiple times. They incorporate random effects parameters into models in addition to fixed effect terms. Random effects parameters account for heterogeneous data with random variability (e.g., both intra and inter-individual). As a result, mixed effects models allow stronger statistical conclusions to be made about correlated observations. The RE-EM tree is an extension of the regression tree for longitudinal data. RE-EM tree has been constructed through an iterative two-step process. In the first step, the random effects are estimated and in the second step, a regression tree is constructed ignoring the longitudinal structure. These two steps are repeated until the random effect estimates converge in the first step. MERF algorithm follows a similar approach to the RE-EM tree, except instead of a decision tree, a random forest is constructed in the second step to achieve improved prediction accuracy and address instability that often plagues a single tree [Briemann,2001]. Further details of these algorithms can be found in Supplementary Material Section B.

**Prediction model performance evaluation:**

Mean absolute error (MAE) was used to evaluate the performance of our models. The Error was defined as the difference between a predicted depression score obtained from the models and the real depression score. MAE is the average of absolute errors.

We validated our model using the cross-validation (Efron & Go 1983) technique, which is being used for the statistical generalizability of a trained model on an independent data set. The main objective of all cross-validation schemes is to ensure that the evaluation step is performed with absolutely no bias and in a fair condition. We have evaluated the model using 10-fold cross-validation (See Supplementary Materials Section C).

## Results

185 participants aged between 20 and 50 (M=21.97, SD=5.97) completed a set of self-report questionnaires at baseline, and at four follow-up assessments. The outcome measure BDI is continuous and spaced three weeks apart. BDI was measured for all 187 participants at baseline followed by 150 on the second visit,137 on the third visit, 122 on the fourth visit, and 113 remaining on the last visit. Descriptive statistics for all predictor variables are summarized in Table 3. Missing values for predictors were not imputed as the proportions of missing values are negligible.

The performances of all models (LMM, RE-EM tree, and MERF) were evaluated using mean absolute error (MAE) and log-likelihood values. The models were validated by a 10-fold CV approach.

**Table 3:** Descriptive Statistics for the predictor variables used in the model.

| Variables | Range | Mean | Standard Deviation |
|---|---|---|---|
| BDI | 0-46 | 6.90 | 7.28 |
| Age | 18-50 | 22.05 | 5.80 |
| CFIPerceivedAlternatives | 0-84 | 64.65 | 10.46 |
| CFIPerceivedControl | 15-52 | 37.47 | 7.04 |
| ProblemFocusedCopingFlexibility | 0-3.0 | .78 | .42 |
| EmotionFocusedCopingFlexibility | 0.0-2.0 | .47 | .36 |
| Reappraisal | 0-42.0 | 29.39 | 7.32 |
| Suppression | 0-28.0 | 14.68 | 5.48 |
| NegativeLifeEvents | 0-43.0 | 8.64 | 6.87 |
| Worry | 16-80.0 | 41.95 | 13.62 |
| Brooding | 5.0-20 | 10.22 | 3.78 |
| Pondering | 5.0-20 | 10.54 | 4.06 |
| CopingFlexibility | 0-1 | .54 | .27 |
| NegativeCognitiveStyles | 1-6.55 | 3.34 | .92 |

The linear mixed-effects model was fitted to predict depression, in which fixed effects were estimated for the demographic and person-centered time-varying predictor (Table3) variables, with a random effect for intercepts. The analysis was conducted using R (R Development Core Team, 2009), with the lme4 package for LMM (Bates et al. 2015). Several interaction terms were included in the model to get the advantage of the fullest model.

RE-EM tree analysis was conducted using the R package REEMtree (Sela & Simonoff, 2015) and the model was fit with autocorrelation after performing log-likelihood testing for autocorrelation. The tree building is based on the R function rpart. The algorithm splits a node where it maximizes the reduction in the sum of squares for the node. Such recursive splitting continues as long as the proportion of variability accounted for by the tree (called the complexity parameter, cp) increases at least 0.001 and the number of observations in the candidate splitting node is greater than 20. After the initial tree is built, it is pruned back based on 10-fold cross-validation. First, the algorithm obtains the tree with final splits corresponding to the cp value with minimized 10-fold cross-validation error. Then, the tree with a final split corresponding to the largest cp value that has a 10-fold cross-validation error less than one standard error above the minimized value is determined as the final tree (the so-called ''one-SE'' rule).
Mixed effects random forest (MERF) was implemented using open-source Python code and the hyperparameters were tuned using grid search. The parameters tuned were the number of trees, the maximum depth of the tree, and the number of iterations. All the features were used to tune MERF's parameters. In MERF the bootstrap sample is drawn on the observation level and predictions are based on the out-of-bag sample for avoiding the risk of overfitting.

**Model Performances using 10-fold schemes.**

We compared mean absolute error and log-likelihood values of LMM, RE-EM trees, and MERF using 10-fold cross-validation. Log-likelihood value is a measure of goodness of fit for any model. The higher the value, the better the model is. Consequently, MAE does not indicate underperformance or overperformance of the model (whether the model under or overshoots actual data). A small MAE suggests the model is great at prediction, while a large MAE suggests that your model may have trouble in certain areas. The MAE reported here is the average error across all observations in the testing set. The MERF model using 10-fold CV performed better with fit (log-likelihood: -1475) and compared to the RE-EM tree (log-likelihood: −1652) and LMM (log-likelihood: -1689). For 10-fold CV MAE values are 3.22, 3.19, and 3.13 for LMM, RE-EM trees, and MERF algorithm, the values are small given that the range for BDI is between 0 and 46. The improvements in predictive accuracy for MERF and RE-EM tree over LMM are 3% and 1% respectively. It indicates that the predictive power of MERF and RE-EM trees is comparable to LMM.

**Significant predictors and interaction between predictors predicting concurrent depression severity.**

The traditional linear mixed effects model yielded eight statistically significant predictors of depression symptoms (Table 4). The model was fitted to predict depression severity measures, in which fixed effects were estimated for age and all measures related to flexibility, cognitive style, emotion regulation, and exposure to negative life events. The correlated nature of observations was accounted for through the estimation of random effects on subjects and time. We found that depression severity was increased significantly when an individual had elevated measures of negative life events, brooding, negative cognitive responses, and interaction between Brooding and negative life events. In contrast, depression severity is decreased significantly when an individual has elevated measures of problem-focused coping flexibility, reappraisal, and interaction between pondering and negative life events.

**Table 4:** Fixed effects for MLM analysis with BDI as an outcome.

| Predictors | b | se | t | p |
|---|---|---|---|---|
| Sex | 2.08 | 0.98 | 2.13 | 0.03 |
| Age | 0.008 | 0.086 | 0.1 | 0.94 |
| CFIPerceivedAlternatives | -0.06 | 0.03 | -1.88 | 0.06 |
| CFIPerceivedControl | -0.1 | 0.05 | -1.62 | 0.1 |
| ProblemFocusedCopingFlexibility | -1.49 | 0.62 | -2.4 | 0.01 |
| EmotionFocusedCopingFlexibility | -0.19 | 0.67 | -0.61 | 0.54 |
| CopingFlexibility | -0.11 | 0.69 | -0.23 | 0.82 |
| NegativeCognitiveStyles | 1.41 | 0.27 | 5.17 | <.001 |
| Reappraisal | -0.14 | 0.04 | -3.39 | 0.001 |
| Suppression | 0.12 | 0.06 | 1.37 | 0.17 |
| NegativeLifeEvents | 0.42 | 0.05 | 7.75 | <.001 |
| Worry | 0.01 | 0.02 | 0.83 | 0.41 |
| Brooding | 0.26 | 0.09 | 3.17 | 0.002 |
| Pondering | 0.15 | 0.09 | 1.5 | 0.13 |
| Brooding * NegativeLifeEvents | 0.08 | 0.03 | 2.69 | 0.01 |
| Pondering * NegativeLifeEvents | -0.07 | 0.03 | -2.4 | 0.02 |

Notes. *b* = unstandardized coefficient, *se* = standard error, *t* = t value for significance, *p* = significance level. All predictor variables were entered at the same time and time-varying predictors were centered around each person's own mean in the model. Only significant interactions are shown here.

The RE-EM tree for depression severity represents relational rules and outcomes presented in Figure 1. The tree was generated using all data. This output provided the additional benefit of conveying relevant clinical information about the model by indicating how a set of influential variables predict depression severity. The terminal nodes represented depression severity. Depression symptoms were best split by brooding, with subsequent groupings based on negative life events and negative cognitive responses. In the lower brooding group (Brooding <13), individuals whose negative life events were greater than or equal to 14 had higher average depression severity (average BDI score: 9.6) than individuals whose negative life events were less than 14 (average BDI score: 5.2). Further down to the leaf node, those with cognitive flexibility control scores less than 36 had less depression than those with cognitive flexibility control greater than or equal to 36.

In contrast, in the group whose brooding was above the initial cut point (brooding ≥ 13), who also had negative cognitive styles greater than or equal to 4 and cognitive flexibility control scores less than 26 had the highest depression severity (average BDI score: 25).

In summary, the RE-EM tree results identified two subgroups of individuals who were best divided based on their brooding score. People in the lower brooding group who were at the lowest risk for depression had fewer negative life events and higher perceived control. People in the higher brooding group had the highest severity of depression if they had more negative cognitive styles and less perceived control. These results highlight the utility of this technique for identifying which predictors are most important, for determining which subgroups are most (and least) likely to have problems with depression, and for identifying which risk factors may be most important to target with interventions to reduce risk.

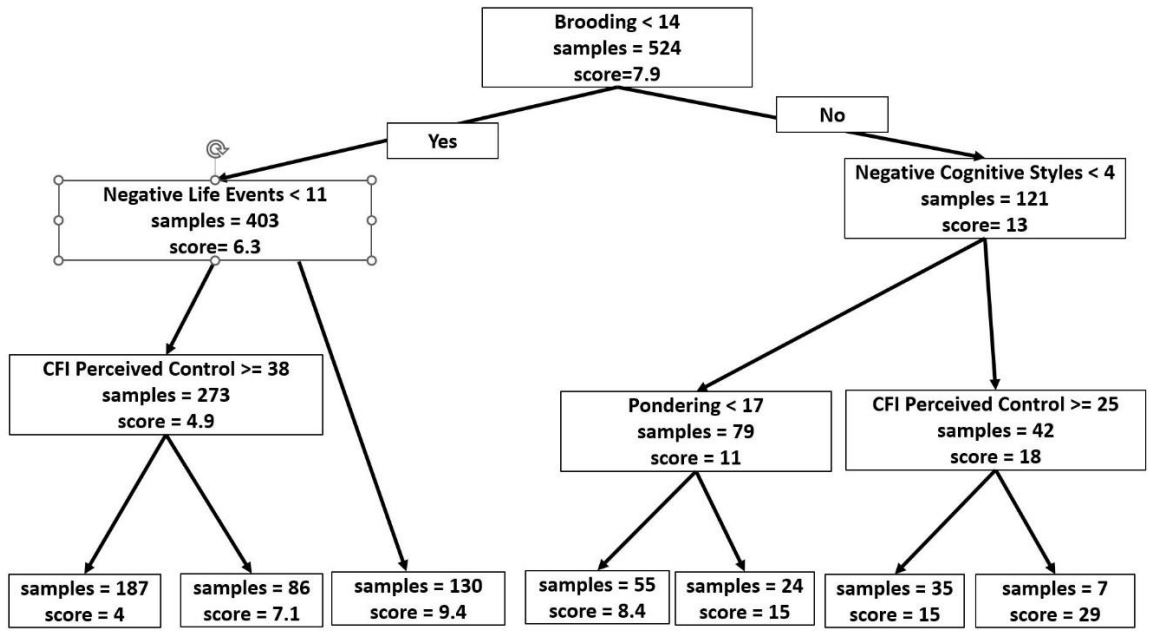

**Figure 1: The structure of the RE-EM tree.** The higher a variable appears in the tree structure, the more predictively significant the variable will be. The ruminative response scale variable 'Brooding' appears at the initial node of the tree as the predictively most important variable. The terminal nodes represented depression severity. The score represents the mean value of the depression scores and the samples represent the number of observations falling in the branch with percentages.

Next, a mixed effect random forest (MERF) model was fitted with all predictor variables, to produce highly accurate prediction due to its robustness. As a random forest algorithm creates many trees, it becomes difficult to visualize all trees. We showed the tree with the best parameter selection (number of trees=300, number of iterations=100, and maximum depth=3) inside the cross-validation procedure. Figure 2 showed that Depression symptoms were best split by negative life events, with subsequent groupings based on brooding and pondering. Other important variables are cognitive flexibility and emotion regulation questionnaires.

People in the higher negative life events group had the highest severity of depression if they had more ruminative responses and less cognitive flexibility. Similar behaviors are observed in people who have fewer negative life events.

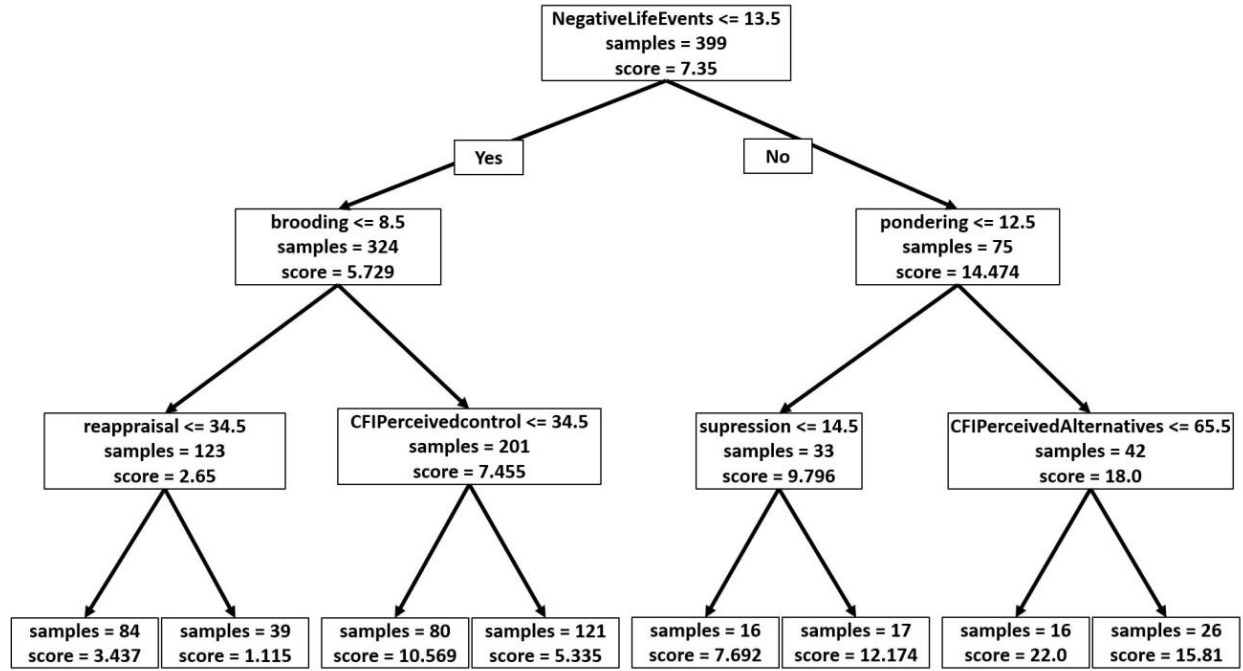

**Figure 2: MERF tree with best parameter selection.** The terminal nodes represented depression severity. The score represents the mean value of the depression scores and the samples represent the number of observations falling in the branch with percentages.

**Discussion**

In this paper, we demonstrated two machine-learning approaches for developing tree models in the presence of multilevel data. The two such methods presented here are the RE-EM tree and MERF. We compared the results with traditional LMM, and our study found that these ML methods performed better in predictive accuracy than LMM. The improvements in predictive accuracy for MERF and RE-EM tree over LMM are 3% and 1% respectively in line with earlier studies comparing mixed-effects decision tree algorithms and traditional mixed-effects algorithms (Sela & Simonoff, 2012, Hajjem et al., 2017, Fokkema et al. 2018). LMM analysis found measures of negative life events, brooding, negative cognitive responses, cognitive flexibilities, problem-focused coping flexibility, and reappraisal to be associated with depression severity. RE-EM tree identified brooding as a root node – that is, the most important initial factor for classification – and negative life events, perceived control, negative cognitive styles, and pondering as internal nodes. In MERF, negative cognitive styles ranked highly, as well as worry, coping flexibility, negative life events, reappraisal, and suppression. The main advantage of the RE-EM tree and MERF over LMM is they can handle large data sets with many covariates, they are robust to outliers and collinearity problems, and they rank the covariates and detect automatically potential interactions between covariates.

One of the advantages of the RE-EM tree and MERF algorithms is identifying and interpreting the relationship between the predictor variables and outcome variables by automatic detection of interactions. For example, using RE-EM trees to predict future depression severity (Figures 4-6) showed that individuals at the highest risk for more severe depression symptoms were those with high brooding (scores > 14) who also had high levels of negative cognitive style (scores > 4), particularly those with less perceived control (scores < 25). Therefore, for individuals with higher levels of brooding, the variables key to determining depression risk appear to be negative life events, negative cognitive style and perceived control. This finding supports the cognitive catalyst model of depression, which proposes that the effects of negative thinking on depression may be amplified by perseverative thinking processes (Ciesla et al. 2007, 2011, Stange et al. 2013,2015). In contrast, individuals with low brooding who had fewer than eleven life events, particularly those with higher perceived control, had the lowest risk for future symptoms of depression. This finding aligns with traditional vulnerability-stress models of rumination as a diathesis for depression (Stange et al. 2014, Abela et al. 2011). This example highlights one of the key ways that tree-based machine learning methods have utility: *they can identify subgroups of individuals (e.g., those with higher or lower levels of brooding) for whom the key variables indicative of risk differ.*

This kind of relationship is hard to obtain in traditional LMM without specifying (a priori) the interaction terms in the model. Thus, RE-EM or MERF has an advantage, particularly for data-driven approaches in which interactions are not specified in advance. Complex interactions in MERF are not easily accessible as it yields the results in a "black box" way, such that the individual trees are not visible. However, based on parameters selected by tuning the tree MERF provides the researcher with an index of variable importance, such that the relative magnitude of relationships with the outcome variable can be compared. Thus, if the researcher would like to understand the specific mechanisms that differentiate individuals on the response variable concerning the predictor variables, then the RE-EM tree may be preferable.

Furthermore, the RE-EM tree directly shows how the relevant patient characteristics should be combined to decide whether a patient is at risk for depression severity. It provides an interpretable summary figure which may assist in monitoring a health behavior intervention. In clinical practice, the fitted RE-EM trees could be used for decision makings and to inform policy.

Predicting future outcomes based on historical observations is a long-standing challenge in many scientific areas. We found that the accuracy of predictions depends on the length of historical observations (the size of the rolling window). As the number of repeated observations used for training increases, the RE-EM tree or MERF model that accounts for random effects performs better. The rules from the RE-EM tree showed that brooding, negative life events, and negative cognitive styles were the main predictors of depression for future time points in our sample. However, it is important to note that although we evaluated several potential risks and protective factors for depression, we did not include an inclusive list of all potential factors (e.g., Hammen, 2018), which could be considered in future research. For example, a recent meta-analysis showed that dampening responses to positive affect may predict increases in depression over time (Bean et al., 2022).

**Limitations**

There are several limitations to tree-based methods. For instance, the cut-points for nodes are done automatically by the package rather than by researchers. The significance of interactions cannot be tested compared to regression modeling. If a researcher wants to test an a priori hypothesis or evaluate the statistical significance of interactions, regression modeling is preferable to tree-based methods. There also were limitations to the example data that were used to demonstrate an application of machine learning. Data were collected among students; although college-age students are at heightened risk for depression and anxiety (Duffy et al., 2019), further investigation using clinical samples (e.g., predicting the future onset of depressive episodes) is needed to improve the generalizability of our findings. Next, measures were self-reported, and thus may be valid only to the extent that individuals are willing and able to report accurately (Haeffel & Howard, 2010). Nevertheless, the methods we describe here also can be used to integrate self-report with other modes of data collection in psychiatric research, such as clinical interviews, psychophysiology, and other biological factors (e.g., fMRI).

Early warning of future depression, which is a primary clinical problem addressed in this study, is important for targeting high-risk patients for monitoring and intervention. The data-driven machine learning methods can help predict future depression changes. These models are interpretable, transparent, and easy to deploy in clinical practice. Lagged training and validation structures can be used to further investigate the temporal effects of risk factors over successive patient visits. For example, regularly-schedule computerized assessments could be fed into ML models to predict sessions when patients are at higher-than-average risk of experiencing an increase in symptoms over the subsequent weeks. This information then could be used to direct resources toward individuals at these moments (e.g., additional interventions, such as medication, skills, and booster sessions) to prevent depression relapse. In conclusion, this manuscript demonstrated that random effects machine learning models can take advantage of increasing sample sizes and dependencies between the observations to generate more robust and accurate models.

**Role of funding:**

Data collection was supported by grants to Jonathan P. Stange from the National Institute of Mental Health (F31MH099761), the Association for Psychological Science, the American Psychological Foundation, and the American Psychological Association. Jonathan P. Stange was supported by grant 1K23MH112769-01A1 from NIMH.

**Author Contributions:**

JS designed and collected data for the original study and provided critical revisions to the paper. RB proposed and implemented the algorithm and wrote the paper.

**Acknowledgments:** None